\documentclass{article}

\usepackage[preprint]{neurips_2021} 

\usepackage[utf8]{inputenc} 
\usepackage[T1]{fontenc}    
\usepackage{hyperref}       
\usepackage{url}            
\usepackage{booktabs}       
\usepackage{amsfonts}       
\usepackage{nicefrac}       
\usepackage{microtype}      
\usepackage{xcolor}         

\usepackage{amsmath}
\usepackage{graphicx}
\usepackage[scaled=.85]{FiraSans}
\usepackage[T1]{fontenc}

\usepackage[font=footnotesize,labelfont=bf,labelsep=quad,justification=justified,singlelinecheck=false]{caption}
\usepackage[font=footnotesize,labelfont=bf,labelsep=quad,justification=justified,singlelinecheck=false]{subcaption}

\hypersetup{colorlinks,linkcolor={blue!40!black},citecolor={blue!40!black},urlcolor={blue!40!black}}

\newcommand{\email}[1]{\href{mailto:#1}{\nolinkurl{#1}}}

\newcommand{\shorturl}[1]{\href{https://#1}{\nolinkurl{#1}}}

\usepackage{csvsimple}

\usepackage{etoolbox}
\patchcmd{\toprule}{\heavyrulewidth}{0.5pt}{}{}
\patchcmd{\bottomrule}{\heavyrulewidth}{0.5pt}{}{}

\usepackage{array}
\newcolumntype{t}{>{\ttfamily}r}

\newcommand{\reducespace}{\vspace{0pt}} 
\usepackage{enumitem}
\newlength{\enumindent}
\newcommand{\cnn}{CNN}
\newcommand{\gpu}{GPU}
\newcommand{\toad}{TOAD}
\newcommand{\vgg}{VGG16}

\title{Understanding top-down attention\\using task-oriented ablation design}

\author{
Freddie Bickford Smith\\University of Oxford\\\email{freddie@robots.ox.ac.uk}
\And
Brett D Roads\\University College London\\\email{b.roads@ucl.ac.uk}
\AND
Xiaoliang Luo\\University College London\\\email{xiao.luo.17@ucl.ac.uk}
\And
Bradley C Love\\University College London\\\email{b.love@ucl.ac.uk}
}

\begin{document}
\maketitle
\setcounter{footnote}{0}
\begin{abstract}
Top-down attention allows neural networks, both artificial and biological, to focus on the information most relevant for a given task.
This is known to enhance performance in visual perception.
But it remains unclear how attention brings about its perceptual boost, especially when it comes to naturalistic settings like recognising an object in an everyday scene.
What aspects of a visual task does attention help to deal with?
We aim to answer this with a computational experiment based on a general framework called task-oriented ablation design.
First we define a broad range of visual tasks and identify six factors that underlie task variability.
Then on each task we compare the performance of two neural networks, one with top-down attention and one without.
These comparisons reveal the task-dependence of attention’s perceptual boost, giving a clearer idea of the role attention plays.
Whereas many existing cognitive accounts link attention to stimulus-level variables, such as visual clutter and object scale, we find greater explanatory power in system-level variables that capture the interaction between the model, the distribution of training data and the task format.
This finding suggests a shift in how attention is studied could be fruitful.
We make publicly available our code and results, along with statistics relevant to ImageNet-based experiments beyond this one.
Our contribution serves to support the development of more human-like vision models and the design of more informative machine-learning experiments.
\end{abstract}
\vspace{5pt}
\section{Introduction}
Ada is apparently nowhere to be seen in the crowded town square.
Spotting her among the hundreds of other people seems an impossible task.
But a moment later she sends a message telling you she is wearing her red coat.
Within seconds you process the few dashes of red in the crowd and you spot Ada.
She was hiding in plain sight.

At play here is the phenomenon of top-down attention.
While the raw sensory input, the pattern of light on the retina, does not change, the processing of that input is dynamically modulated so that task-relevant visual features are emphasised \citep{carrasco11,kruschke92,lindsay20a}.
This is useful: it boosts perceptual abilities.
Intriguingly, it is more useful in some tasks than in others.
Suppose now you are trying to spot Ada but the town square is not crowded at all.
This time it is impossible \emph{not} to see her.
Attention cannot make much difference: there is little benefit to emphasising task-relevant information.
Empirical evidence agrees with this intuitive case (Section \ref{sec:background}).

The task-dependence of attention's influence provides a window into issues at the core of attention research.
What aspects of a task does it help to deal with?
If it were absent, in what ways would a task become more challenging?
Cognitive science's answers to these questions remain incomplete despite more than a century of investigation \citep{james1890,helmholtz1896}.
While lab experiments with simple visual stimuli have yielded valuable findings, it remains an open question whether these findings hold in naturalistic settings \citep{peelen14}.
This is a considerable gap in the theory of attention, a gap that matters both for understanding human cognition and for developing new machine-learning methods.
Addressing this requires dealing with the complicated, multivariate nature of real-world visual tasks.
Seeking a tractable means of going about this, we propose characterising \emph{when} attention helps as a way of explaining \emph{how} it helps.
If attention produces a perceptual boost in one task but not in another, the differences between the tasks point to the mechanism at play.

Understanding how the influence of attention varies in naturalistic settings, where sources of variation are complex and numerous, dictates a systematic and large-scale experimental approach.
Computational modelling with a convolutional neural network ({\cnn}) makes this possible by allowing us to scale human-like perception to thousands of visual tasks: although not perfect, {\cnn}s are state-of-the art models of human vision \citep{kriegeskorte15,lindsay20b,serre19,yamins16}.
This approach also allows us to examine phenomena internal to the model, such as the representational similarity of stimuli, not just simple stimulus statistics.
Our experiment thus stands to help develop a richer cognitive account of visual perception.
At the same time, a careful analysis of attention's influence in a {\cnn} serves to inform the development and deployment of machine-learning methods in settings other than cognitive science.

We hypothesise that qualitative differences between visual tasks can be captured in numerical quantities, and that these quantities will help explain why attention is more useful in some tasks than in others.
Each task we consider consists of classifying images from a chosen task set, a pair of classes from the ImageNet dataset \citep{russakovsky15}.
Based on past work as well as intuitions about how our model might behave, we identify six quantitative dimensions along which there is substantial variation between task sets.
We then aim to establish a relationship between these task-set properties and the extent to which attention produces a perceptual boost. 
Does attention's role in our model match findings from lab experiments using simple stimuli, or is it altogether different when we consider more complex settings?
It is unclear at the outset which will be true. 
Our investigation reveals that it is the latter: the influence of attention is strongly connected to \emph{system-level} variables---those associated with the model, the distribution of training data and the task format---rather than \emph{stimulus-level} variables that describe visual appearance. 

\begin{figure}
\centering
\includegraphics[width=\textwidth]{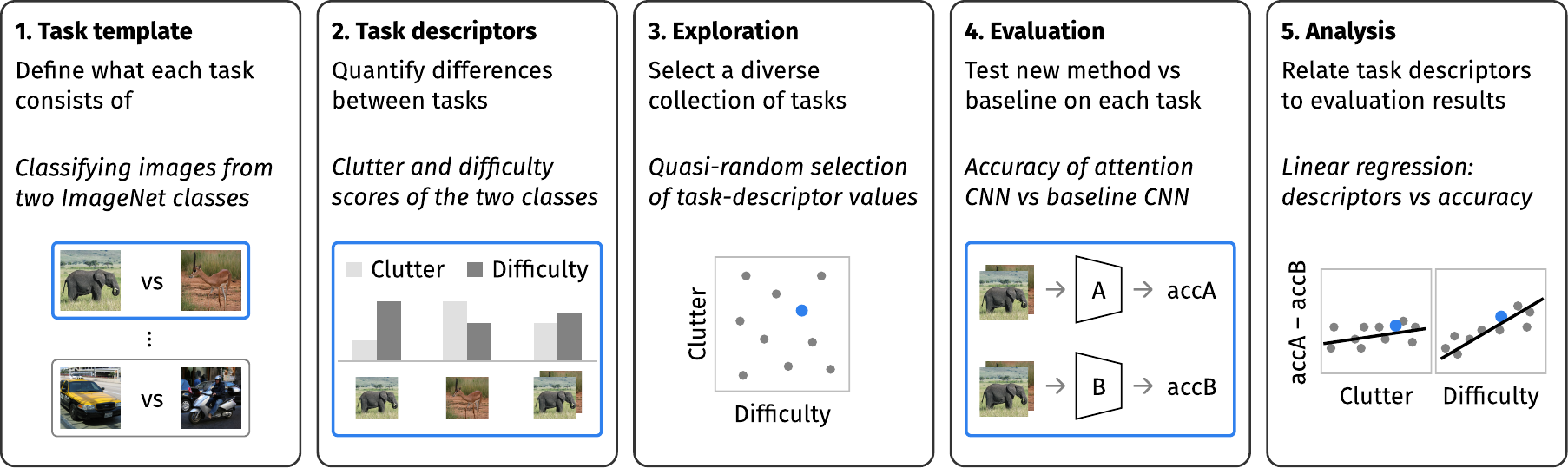}
\caption{
Task-oriented ablation design ({\toad}), a general experimental framework for understanding the interaction between a machine-learning method and the tasks it is applied to.
The core idea is to make quantitative comparisons between tasks, supporting more systematic exploration and more granular analysis than in a typical ablation study, which involves comparing a new method to a baseline method in terms of average performance on a standard test dataset.
{\toad} results in a precise description of the relationship between the nature of a task and the performance impact of a methodological change.
For each step, we show a simplified version of how we do that step in our study of attention, with the highlighted element (in blue) corresponding to a single task within the experiment.
Other possible applications of {\toad} include understanding how the accuracy of a sentiment-analysis technique varies with language style, how the error of a rainfall-forecast model varies with the climate, and how the intersection-over-union metric of a medical-imaging method varies between cancers.
}
\label{fig:toad}
\reducespace
\end{figure}

In the process of testing our hypothesis, we define an experimental framework applicable across a broad range of settings in machine-learning research.
Ablation studies provide crucial insight into the effects of changing an algorithm or model. With task-oriented ablation design ({\toad}), it is possible to extract even more scientific understanding from each ablation study.
The core idea of {\toad} is to make quantitative distinctions between tasks in order to support more targeted experiments and more granular analysis than usual (Figure \ref{fig:toad}).
Despite its simplicity, {\toad} is a powerful framework, allowing detailed insights into the relationship between the nature of a task and the effects of a methodological change.
Our study of attention is a concrete demonstration of this: {\toad} makes it clear how to capitalise on the diversity of ImageNet so as to understand the complex factors underlying the task-dependence of attention's perceptual boost.

\begin{figure}
\centering
\includegraphics[width=\textwidth]{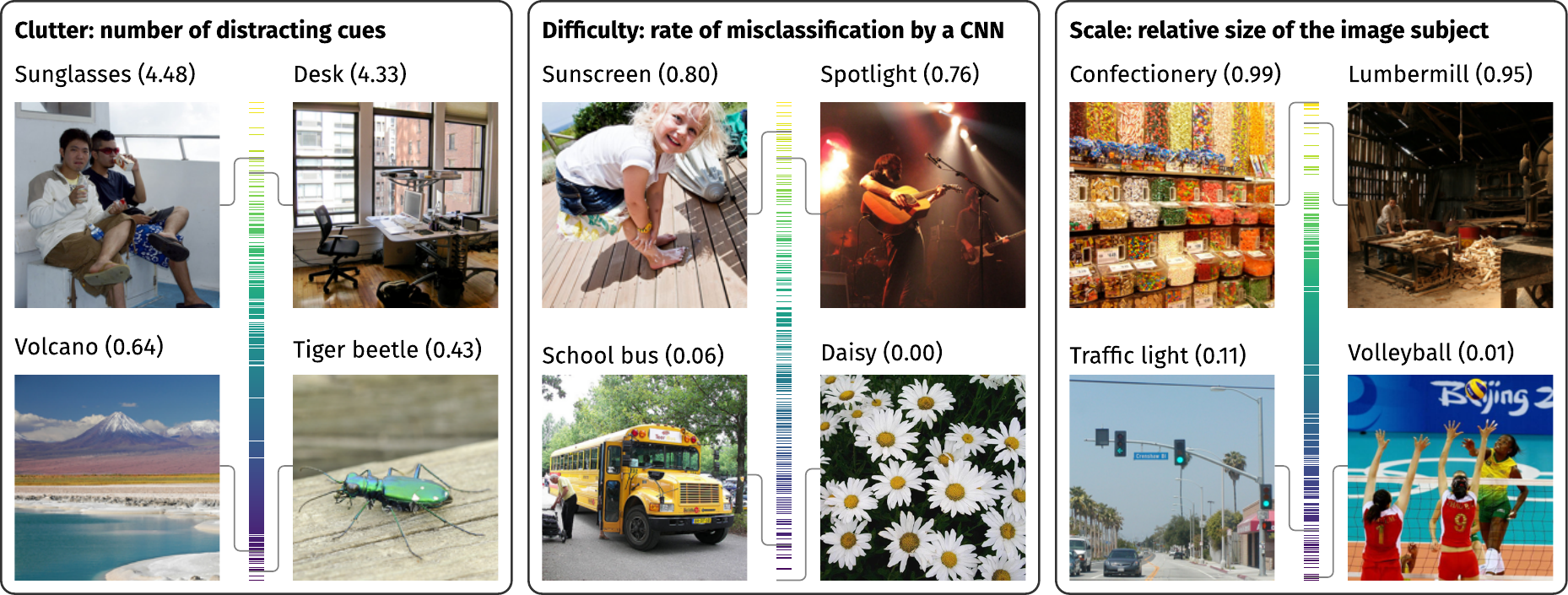}
\caption{
Three class-level statistics that help quantify the diversity of ImageNet.
Clutter represents how much distracting information is present in a typical image from the class.
Images from high-clutter classes often include multiple objects that have little relevance to the subject of the image; those from low-clutter classes have few or no distracting cues.
Difficulty is the misclassification rate of an ImageNet-trained convolutional neural network evaluated on images from the class.
Images from high-difficulty classes are often ambiguous. 
Scale is the relative size of the subject in a typical image from the class.
Images from high-scale classes are almost completely occupied by the subject; those from low-scale classes have the subject appear in only a small proportion of the pixels.
For each statistic, we show the distribution of values across ImageNet (coloured lines; lighter hue indicates higher score) along with representative images from selected classes.
}
\label{fig:clutter_difficulty_scale}
\reducespace
\end{figure}

\begin{figure}
\centering
\includegraphics[width=\textwidth]{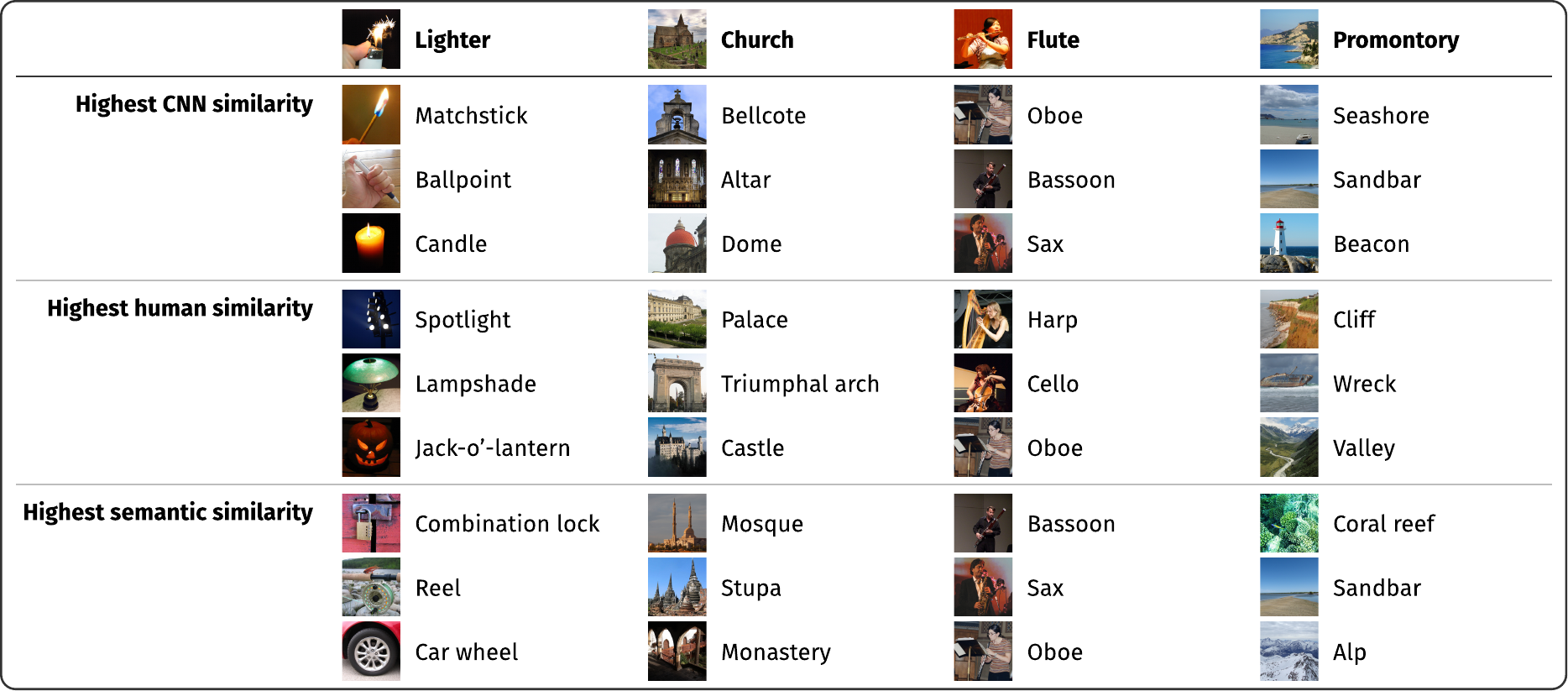}
\caption{
Three ways of measuring the similarity of a pair of ImageNet classes.
{\cnn} similarity reflects how frequently the two classes are confused for each other by an ImageNet-trained convolutional neural network.
Human similarity relates to how closely people associate the two classes, inferred from comparisons they make between images in psychological tests.
Semantic similarity represents the conceptual proximity of the two classes, according to the semantic hierarchy associated with ImageNet.
To contrast the three similarity measures, we show four query classes along with the most similar classes according to each measure.
For some queries, the three class rankings are alike; for others, the rankings are notably different.
The `lighter' query illustrates the source of differences: {\cnn} similarity ranks classes based only on visual appearance; semantic similarity ranks classes based only on abstract concepts; human similarity involves a mixture of the two.
}
\label{fig:similarity_measures}
\reducespace
\end{figure}

As well as defining a broadly applicable experimental framework, we provide data that is similarly transferable to other lines of inquiry.
This data takes the form of statistics describing how classes vary within ImageNet.
For each individual class, we provide three scores: clutter, difficulty and scale (Figure \ref{fig:clutter_difficulty_scale}).
For each pair of classes, we provide three measures of similarity: one based on {\cnn} ouputs, one based on human judgements and one based on a formal semantic hierarchy (Figure \ref{fig:similarity_measures}).

Our contribution is thus threefold.
First, we systematically characterise the perceptual boost of top-down attention in naturalistic visual tasks, considering system-level variables not previously studied in this context.
Our findings contrast with expectations based on prior work, with implications for how models of human visual perception should be analysed and developed.
Second, we define {\toad}, a general framework for quantitatively understanding the interaction between a methodological change and the tasks it is applied to.
This has the potential to see wide use in machine learning.
Third, we provide ready-to-use statistics describing the 1000 classes that comprise ImageNet.
These should help the research community better understand this widely used dataset.
Collectively, the three components of our contribution stand to advance both cognitive science and machine learning.
\section{Background and related work}
\label{sec:background}
Human visual perception is in a constant state of flux: the brain processes an identical retinal stimulus differently whenever the task or context changes \citep{carrasco11}.
A key contributor to this flexibility is attention, which emphasises the visual information most relevant in a given setting.
Our work concerns \emph{top-down} attention, where relevance depends on the task at hand and the surrounding context; this contrasts with bottom-up attention, where relevance is determined solely by the stimulus \citep{borji13,itti01}.
Top-down attention is known to enhance people's performance on visual tasks \citep{lindsay20a}.
At the same time, it remains poorly understood exactly how attention brings about its perceptual boost.
What aspects of a task does attention interact with? In what ways would a task become harder if attention were not applied?
Addressing these questions is crucial for fundamental research in cognitive science but also for transferring ideas to machine-learning methodology \citep{hassabis17}.

\paragraph{Task-dependence as a window into attention's role}
How can we better understand attention's role in enhancing perceptual abilities?
We propose building on the finding that the influence of attention varies between tasks.
Attention's effect on signal-detection sensitivity and bias varies with task type \citep{downing88}; its modulation of visual area 1, visual area 4 and the inferior temporal cortex varies with task difficulty \citep{boudreau06,chen08,spitzer88,spitzer91}; its impact decreases with greater scene complexity \citep{rolls08}; its strength varies with the size of the stimulus relative to the attentional field \citep{lindsay20c,reynolds09}; and its modulation of the middle temporal visual area changes with the number of stimuli in the receptive field \citep{lee10}.
While this is compelling evidence for the task-dependence of attention's influence, it does not necessarily tell the whole story.
There are many different conceptions of attention; the findings cited above are unlikely to all describe the same mechanism.
In addition, existing experimental data was gathered using a narrow range of simplistic visual tasks, which to some degree reflects the practicalities of lab studies in cognitive science \citep{barbosa21}.
There have been few studies of attention's impact across a broad range of naturalistic tasks \citep{peelen14}.

\paragraph{Studying attention in a convolutional neural network}
Convolutional neural networks are state-of-the-art models of human visual perception in terms of both explaining neural activity and predicting behaviour \citep{lindsay20b,schrimpf20}.
Computational modelling with {\cnn}s is thus a way of developing a cognitive account of naturalistic vision at a level of complexity previously impossible due to the practical barriers mentioned above.
It also provides an exciting opportunity to exchange ideas between cognitive science and machine learning \citep{luo21}.
Top-down attention can be modelled as a modulation of the bottom-up representation computed by a standard {\cnn}, analogous to the influence of the prefrontal cortex on the visual cortex \citep{bar03,bichot15,lindsay20a,paneri17,rossi07}.
This shapes the flow of information to emphasise task-relevant features, supporting adaptation between tasks.

Our model is intentionally straightforward and similar to those used in related work.
The foundation of the model is {\vgg}, a widely used {\cnn} that takes as input an image and produces as output a predicted label for the image \citep{simonyan15}.
Following Occam's razor \citep{mackay03}, we use one of the simplest possible attention mechanisms: attention modulates the representation at the final convolutional layer, multiplying each feature map by a separate weight.
As well as being conceptually simple, this model closely resembles the attention-augmented {\cnn}s used by \cite{bickfordsmith20}, \cite{lindsay18} and \cite{luo21}.
This similarity supports straightforward comparisons between past work and ours.
It also allows us to focus on drawing insights from the model instead of worrying about whether the model is useful in the first place.

Three aspects of this model are valuable to note to understand how it fits into the broader literature.
First, the attention weights are a function \emph{only} of the task at hand: if the task does not change, the weights do not change, even if the input to the network does.
This makes the mechanism different from many in machine learning, including those proposed by \cite{ba15}, \cite{chen17}, \cite{devries17}, \cite{denil12}, \cite{dosovitskiy21}, \cite{larochelle10}, \cite{mnih14}, \cite{perez18}, \cite{ranzato14}, \cite{stollenga14} and \cite{vaswani17}, as summarised in a recent review \citep{lindsay20a}.
Second, attention here is feature-based as opposed to spatial: when a representational feature is modulated, it is modulated across the whole visual field.
Attention is also covert, meaning there is no notion of glancing between points in space.
Third, the model is feedforward, giving a setup comparable to short-duration object-recognition episodes in studies of human perception \citep{dicarlo12}.

\paragraph{Using ablation studies to understand machine-learning methods}
Having defined a model of visual processing with attention, the challenge is to understand it.
A simple yet powerful approach is to perform an ablation study.
The general formula for an ablation study is to evaluate the performance of method A relative to method B on a task.
In a typical setup, method A incorporates a design change, method B is a relevant baseline (often a version of method A where the design change is disabled) and evaluation involves measuring average performance across a test dataset.
The relative performance of the methods reveals information about the effects of the design change.
This information is prized in machine learning: ablation studies are the predominant way of assessing new techniques.

While valuable, the typical setup needlessly throws away useful information: comparing \emph{average} performance amounts to treating the test dataset as a black box.
Task-oriented ablation design instead recognises a dataset as a diverse mix of examples, subsets of which can be used to define tasks.
After identifying factors of variation between tasks, it becomes straightforward to derive an experiment that reveals how the nature of a task relates to the effects of a methodological change (Figure \ref{fig:toad}).

{\toad} is at the core of our investigation.
It guides us in capitalising on the diversity of ImageNet: instead of using the dataset to define a single task, we use it to generate a wide array of smaller tasks, across which we can compare the perceptual boost of attention.
It also streamlines execution: having identified candidate tasks and quantified differences between them (steps 1-2 in Figure \ref{fig:toad}), the rest of the process (steps 3-5) can be automated.

Since {\toad} is simple and general by design, its possible applications extend far beyond our experiment. 
In natural-language processing, it could help determine how text style (eg, sentence length) affects the accuracy of a sentiment-analysis technique.
In meteorology, it could be used to understand how the errors made by a rainfall-forecast model depend on the climate (eg, regional average temperature).
In computational medicine, it could reveal how the predictions of an image-segmentation method vary in quality between forms of cancer (eg, growth rate).
Whereas a standard ablation study would give a crude indication of whether a new method works better or worse, applying {\toad} leads to specific insights about what it is about a task that a method does or does not deal with.

\paragraph{Measuring variation across ImageNet-based tasks}
ImageNet is a strikingly diverse dataset: its 1000 classes constitute a rich cross-section of the naturalistic visual world, and \emph{within each class} there is a large degree of heterogeneity \citep{roads21,russakovsky15}.
This makes it an excellent source of data for an experiment based on the {\toad} framework.
In our experiment, each task consists of classifying images from a task set comprised of two ImageNet classes.
To measure how tasks differ, we identify six task-set properties (factors of variation) that might be linked to the influence of attention in our model: clutter, difficulty, scale, {\cnn} similarity, human similarity and semantic similarity.
Notably, these factors are likely to be relevant to other ImageNet-based studies.
\section{Method}
\label{sec:method}
How does the perceptual boost of top-down attention vary with the nature of a visual task?
To answer this, we use an instance of task-oriented ablation design.
The experiment consists of five parts.
\begin{enumerate}[leftmargin=\enumindent,itemsep=0pt]
\item Task template: classifying images from a task set comprised of two ImageNet classes.
\item Task descriptors: six task-set properties based on class-level statistics.
\item Exploration: quasi-random selection of tasks using a Sobol sequence.
\item Evaluation: comparing the accuracy of two {\cnn}s, one with top-down attention and one without.
\item Analysis: linear regression with the task-set properties as covariates and accuracy as response.
\end{enumerate}
Code to reproduce our experiment, along with associated data, is available at \shorturl{github.com/fbickfordsmith/attention-toad} (MIT license).
See Appendix \ref{sec:resources} for details on the resources used in the experiment. 

\subsection{Tasks: image classification on pairs of ImageNet classes}
\label{sec:tasks}
Each task we consider consists of classifying images from a chosen \emph{task set}, a pair of ImageNet classes.
There can be clear qualitative differences between task sets: compare, say, \{magpie, robin\} to \{magpie, ambulance\}.
For precision and scalability, we aim to capture such differences with six numerical quantities called \emph{task-set properties}.
Then, to ensure we consider a broad range of tasks, we select a collection of task sets that vary widely with respect to these properties.
We cover the key ideas here and provide a more detailed description in Appendix \ref{sec:task_set_properties}.

\paragraph{Task-set properties}
Clutter represents how much distracting information is present in a typical image from a task set.
To measure this, we use an object detector to count the number of distracting cues in each image, building on work by \cite{kuznetsova20}, \cite{ren15}, \cite{russakovsky15} and \cite{szegedy17}.
Difficulty is the error rate of our baseline model (Section \ref{sec:model}) on images from a task set.
Scale is the relative size of the subject in a typical image from a task set.
We estimate this using bounding-box annotations associated with ImageNet.
{\cnn} similarity reflects how frequently the classes in a task set are confused for each other by our baseline model.
Human similarity relates to how closely people associate the classes in a task set, inferred from comparisons they make between images.
For this measure, we make use of psychological embeddings developed by \cite{roads21}.
Semantic similarity represents the conceptual proximity of the classes in a task set according to WordNet \citep{miller95}, the semantic hierarchy associated with ImageNet.
Computing the pairwise similarity of classes is made straightforward by semantic embeddings created by \cite{barz19}.

\paragraph{Selecting task sets}
Having defined six task-set properties, we need to select a collection of task sets that covers an interesting range of combinations of these properties.
To do this, we enumerate all possible task sets and compute their properties.
This gives an interval of possible values for each property.
The Cartesian product of these intervals is a volume in six-dimensional space.
Systematically exploring combinations of the task-set properties requires filling this volume evenly with a collection of points, where each point represents the properties of a task set.
An effective way of doing this is to generate points with a Sobol sequence, a low-discrepancy sequence sometimes used for quasi-Monte Carlo integration \citep{morokoff95}.
In general, these points correspond to hypothetical, not actual, task sets.
So we treat them as targets and find actual task sets whose properties are closest to the targets (Figure \ref{fig:task_set_properties}, Appendix \ref{sec:task_set_properties}).

\subsection{Model: attention-augmented convolutional neural network}
\label{sec:model}
\vgg, the foundation of our model, is a function, $f$, that takes an image, $x \in \mathbb{R}^{224 \times 224 \times 3}$, as input and predicts the ImageNet class, $c \in \{1,2,3,\ldots,1000\}$, the image belongs to:
\begin{equation*}
f(x) = p(c \mid x)
\end{equation*}
This function can be decomposed into two parts.
The convolutional layers, $f_c$, transform $x$ to a latent representation, $z \in \mathbb{R}^{7 \times 7 \times 512}$.
The densely connected layers, $f_d$, map from $z$ to $p(c \mid x)$.
That is,
\begin{equation*}
f_c(x) = z \qquad \qquad f_d(z) = p(c \mid x)
\end{equation*}
In our model, top-down attention is a multiplicative modulation of $f_c(x) = z$ by a collection of nonnegative attention weights, $a \in \mathbb{R}_{\geq 0}^{512}$.
The result is an augmented version of {\vgg} given by
\begin{equation*}
\hat{f}(x,a) = p(c \mid x,a) = f_d(a \odot z) \qquad \qquad (a \odot z)_{i,j,k} = a_k z_{i,j,k} 
\end{equation*}

\paragraph{Training}
We treat the ImageNet-pretrained $f_c$ and $f_d$ as fixed functions: the attention weights are the only trainable parameters.
Training consists of minimising the cross entropy between the model's predictions and the dataset labels.
Each update to the attention weights is computed by the Adam optimiser \citep{kingma15} using gradients computed on a minibatch of 128 examples (Adam settings: $\alpha=0.0003$, $\beta_1=0.9$, $\beta_2=0.999$, $\hat{\varepsilon}=10^{-7}$).
Images are preprocessed in the same way as for {\vgg}.
Training stops once the cross entropy starts to increase on the validation dataset.

\paragraph{Baseline}
To set up the baseline model for our experiment, we set the value of each attention weight to 1 and then train the weights on examples from all 1000 ImageNet classes.
The resulting weights are task-generic: they are optimised for ImageNet as a whole, not for a particular task set.
Using an initial value of 1 for the weights means that at the beginning of training the network has the same input-output mapping as a standard {\vgg}.
The heterogeneity of {\vgg}'s pretrained parameters means that homogeneity in the attention weights does not cause training issues.

\paragraph{Top-down attention}
To produce attention weights specialised for a task set, we initialise the weights to those of the baseline model and then train them on a mixture of examples from both inside and outside the task set.
In order to balance the performance costs and benefits of attention \citep{luo21}, we include equal numbers of task-set and non-task-set examples, with a fresh random sample of non-task-set examples for each epoch.
For instance, suppose the two classes that comprise a task set each have 1000 examples in the training dataset.
When training attention weights on this task set, updates are computed using all 2000 examples from the task set as well as 2000 examples from the 998 ImageNet classes not in the task set.
The non-task-set examples are drawn evenly from the classes: in this case, there are at least two examples from each of the 998 classes.

\paragraph{Evaluation}
Inferring the perceptual boost of attention on a task set is straightforward.
We retrieve the ImageNet validation images that belong to the classes in the task set.
Then, using those images, we assess the accuracy of the baseline model as well as the accuracy of a model whose attention weights are optimised for the task set.
The difference between those two accuracies indicates the impact of applying task-specific attention weights.
\section{Results}
\label{sec:results}
We present a large-scale source of empirical evidence on the task-dependence of attention's influence in naturalistic vision.
This is the product of training attention weights on 2000 task sets, requiring thousands of hours of {\gpu} runtime.

\begin{figure}
\centering
\begin{subfigure}{0.46\textwidth}
    \includegraphics[width=\textwidth,trim={0.3cm -0.3cm 0.5cm 0}]{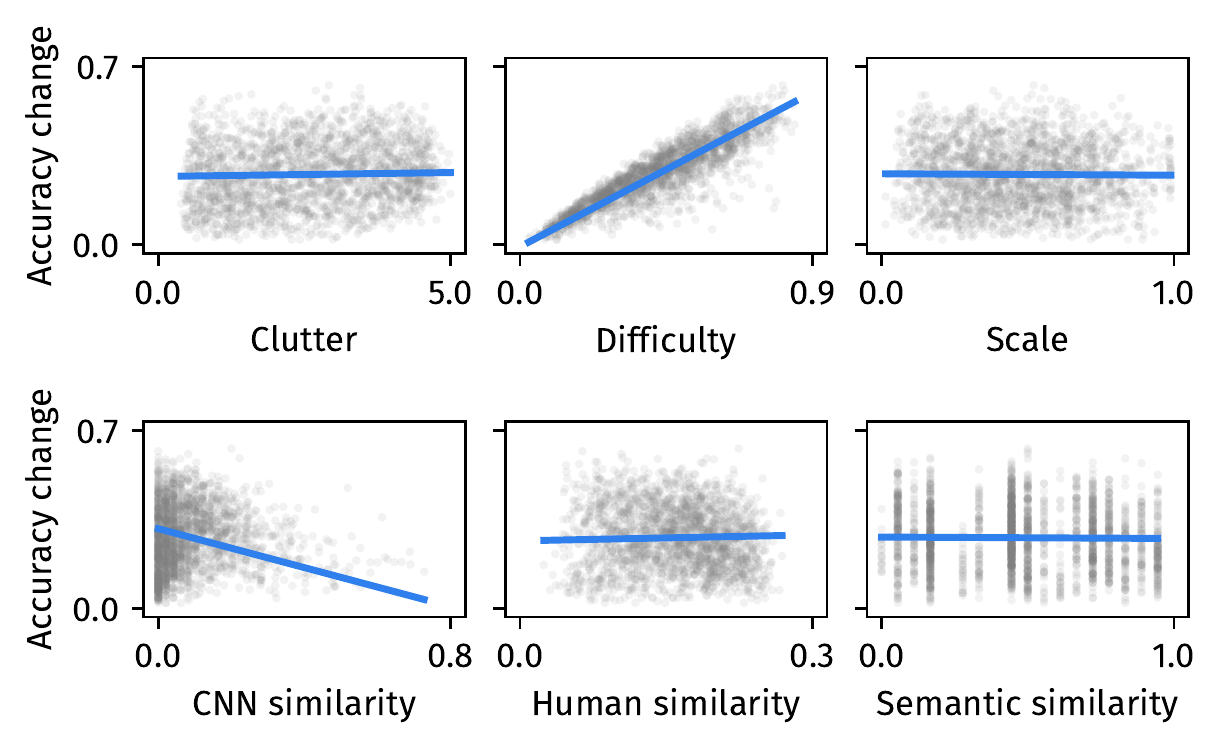}
\end{subfigure}
\hfill
\begin{subfigure}{0.48\textwidth}
    \textsf{\scriptsize
    \begin{filecontents*}{figures/accuracy_change.csv}
i,beta_i,beta_hat_i,stderr_beta_i,variable
0,0.013,0.277,0.005,Intercept
1,0.003,0.004,0.001,Clutter
2,0.675,0.120,0.006,Difficulty
3,-0.006,-0.001,0.005,Scale
4,-0.383,-0.042,0.015,CNN similarity
5,0.078,0.004,0.023,Human similarity
6,-0.006,-0.002,0.004,Semantic similarity
\end{filecontents*}
\csvreader[
        head to column names,
        tabular=ttttl,
        table head=\toprule $i$ & $\beta_i$ & $\beta_i'$ & $\textrm{SE}(\beta_i)$ & Description \\ \midrule,
        table foot=\bottomrule,
        ]{figures/accuracy_change.csv}{}{\csvlinetotablerow}
    }
    \caption*{\sffamily \scriptsize
    Coefficients of the linear-regression model shown in the plots, for which $R^2=0.88$.
    $\beta_i'$ is the standardised version of $\beta_i$.
    $\textrm{SE}(\beta_i)$ is the standard error of $\beta_i$.
    For $i \in \{1, 2, 4, 5\}$, $p(\beta_i)<0.005$.
    }
\end{subfigure}
\vspace{-0.2cm}
\caption{
The perceptual boost produced by top-down attention on 2000 visual tasks.
Each task consists of classifying images from a task set (pair of ImageNet classes).
The perceptual boost of attention is the accuracy change when switching from task-generic attention weights (baseline) to task-specific attention weights in a convolutional neural network.
The strength of the perceptual boost varies with the properties of the task set: clutter, difficulty, {\cnn} similarity and human similarity all have statistically significant effects.
Each grey point corresponds to a single task set.
The blue lines illustrate a linear-regression model fit by least-squares estimation.
}
\label{fig:accuracy_change}
\reducespace
\end{figure}

\paragraph{Relating task-set properties to attention's perceptual boost}
For all task sets, applying top-down attention leads to an improvement in the model's image-classification accuracy.
This perceptual boost is stronger when clutter increases, difficulty increases, {\cnn} similarity decreases, or human similarity increases (Figure \ref{fig:accuracy_change}).
Scale and semantic similarity do not have statistically significant effects.

\paragraph{Explaining the significance of difficulty}
Difficulty is the task-set property most closely associated with the perceptual boost of attention.
Aiming to understand this, we compare the task-generic attention weights of the baseline model to the task-specific attention weights trained on each of the task sets in the experiment (Figure \ref{fig:attention_weights}).
This yields a key insight into what underlies the difficulty of a task set and, in turn, why attention's perceptual boost is greater on more difficult task sets (Section \ref{sec:discussion}).

\paragraph{Exploring within-class variation in clutter and scale}
Clutter and scale vary substantially between images belonging to the same class (Figure \ref{fig:within_class_variation}, Appendix \ref{sec:task_set_properties}).
This observation supports an explanation for why clutter and scale have relatively little effect on the perceptual boost of attention (Section \ref{sec:discussion}).
\section{Discussion}
\label{sec:discussion}
It is striking that stimulus-level variables traditionally associated with attention, such as visual clutter and object scale, have less explanatory power than system-level variables that capture the interaction between the model, the distribution of training data and the task format.
This implies tension between our results and expectations based on prior studies involving simple stimuli.
One interpretation of this finding is that understanding attention's influence in our model requires resisting simplistic, stimulus-level descriptions of task differences.
Consider difficulty, the task-set property most strongly linked to the influence of attention.
It is common to frame difficulty as an inherent property of particular images.
This is certainly true to some extent: some images portray an object unambiguously while others do not (compare `daisy' to `sunscreen' in Figure \ref{fig:clutter_difficulty_scale}).
But there is more going on than just this.
In our experiment, even a simple attention mechanism yields substantially improved performance on the most difficult task sets (Figure \ref{fig:accuracy_change}).

\begin{figure}
\centering
\includegraphics[width=0.7\textwidth,trim={0 0.2cm 0 0}]{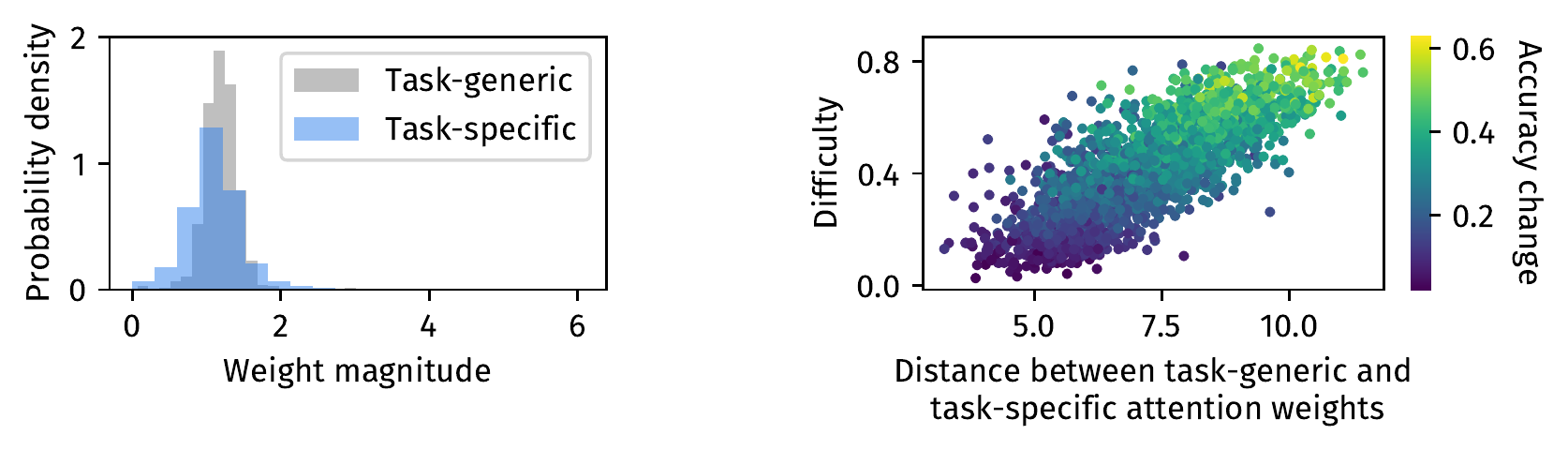}
\caption{
Comparing task-generic and task-specific attention weights.
Left: the distribution of task-specific weights is broader than the distribution of task-generic weights, with standard deviations of 0.39 and 0.23 respectively.
Right: the shift from task-generic to task-specific weights is larger for more difficult task sets, with a correspondingly larger perceptual boost of attention.
For each task set, we plot (a) the Euclidean distance between the attention weights trained on the task set and the task-generic attention weights of the baseline model against (b) the difficulty of the associated task set.
The colour of each point denotes the accuracy change produced by switching from the task-generic weights to the task-specific weights.
}
\label{fig:attention_weights}
\reducespace
\end{figure}

Analysing the attention weights of our model reveals why attention is so useful for difficult tasks.
Training these weights (and the parameters of neural networks more generally) involves trading off features in terms of their overall contributions to performance \citep{hermann20a,sutton06}.
A feature might be discriminative for some examples but confounding for others; the weight placed on that feature is determined by the feature's overall effect on performance across the training data.
The baseline version of our model has task-generic attention weights that trade off features such that the model's performance is (locally) optimal when evaluated across all ImageNet classes.
Training the model on a particular task set, comprising only two ImageNet classes, corresponds to a relaxation of this feature tradeoff.
The attention weights can now prioritise features that are helpful for this task but were previously suppressed.
As a result, the distribution of attention weights changes; crucially, it changes more drastically as difficulty increases (Figure \ref{fig:attention_weights}).
This reveals a deep insight into what makes a task set difficult.
The most difficult task sets are those for which the optimal set of attention weights are most dissimilar from the baseline set of attention weights; they are also the task sets for which attention produces the greatest change in the model's accuracy.
In other words, attention has greater impact for more difficult task sets because their key discriminative features are more strongly suppressed when training on all ImageNet classes.
From this perspective, difficulty is clearly a system-level phenomenon rather than just an inherent property of a particular image.
It encompasses the host of factors that determine the representations the model learns to compute and how those representations are weighted.

A similar argument justifies the importance of {\cnn} similarity relative to human similarity and semantic similarity.
Whereas the latter two measures are derived from external data sources, {\cnn} similarity incorporates the complex interplay between the model and external factors. 
Taking an internal view provides an intuitive case for the link between {\cnn} similarity and the perceptual boost of attention.
For attention to help discern between the two classes in a task set, it must weight the features of the model's representations such that the dissimilarity of the two classes is emphasised, leading to separability at the model's output.
If the two classes in a task set have low {\cnn} similarity, the classes' internal representations are unalike, which makes it straightforward to find attention weights that emphasise the dissimilarity.
If the two classes have high similarity, the opposite is true.

Why is clutter, the presence of distracting visual cues, not strongly linked to the perceptual boost of attention in our model?
Clutter is often cited in descriptions of attention \citep{chikkerur10,navalpakkam07,peelen14}.
The association is a natural one to make: an empirically validated theory is that stimuli compete for limited computational resources \citep{carrasco11}; viewing attention as a mechanism for allocating those resources \citep{lindsay20a}, if follows that clutter and attention should be linked.
Yet our results do not support this.
The variation of clutter across the tasks we study barely affects the influence of attention.
One possible explanation is that ImageNet-trained {\cnn}s process images in a less object-centric way than humans do.
{\cnn}s recognise textures more than they do shapes \citep{baker18,geirhos19,hermann20b,kucker19}.
This means that the model we use is relatively insensitive to clutter by default, which in turn means that attention does not have much impact.
This contrasts with attention in humans \citep{nuthmann10}.
If future work aims to replicate in models the influence of attention as measured in lab studies, it might be necessary to address the texture bias of models.

Scale, the relative size of the subject in a typical image, also has little effect on the perceptual boost of attention.
It is tempting to argue that scale is a spatial phenomenon and is thus unlikely to interact with the feature-based attention mechanism in our model.
But this argument is probably too simplistic.
Scale is closely linked to what visual features are present in an image: even for a fixed image subject, changing the scale can result in considerable differences in terms of visual features.
A more convincing explanation is that scale is an instance-level phenomenon that needs to be dealt with using bottom-up signals.
There is so much within-class variation in scale (Figure \ref{fig:within_class_variation}, Appendix \ref{sec:task_set_properties}) that it is hard to determine a set of attention weights that is useful on average.
In this light, it seems sensible that the top-down form of attention in our model, in which attention weights are purposefully not conditioned on image information, has little interaction with scale.
The same logic provides further grounds for explaining the clutter result discussed above: the task-level attention weights do not deal with clutter because it varies so much from one image to another.

\section{Conclusion}
What does top-down attention do to enhance performance in naturalistic visual tasks?
To understand \emph{how} attention helps, it is instructive to work out \emph{when} it helps.
Addressing this at a level of scale and complexity beyond that in past studies, we investigate the perceptual boost of attention in a {\cnn} on thousands of tasks.
Task-oriented ablation design, a broadly applicable experimental framework, makes this possible.
Our key finding is that explaining the influence of attention in our model demands a system-level perspective on what makes a task challenging and how attention helps to deal with this.
This raises the question of whether, in the context of real-world visual tasks, it is actually helpful to observe the effect of a small number of simple stimulus-level variables.
Instead, perhaps we must take an inside view, informed by variables that require access to the model and the data it is trained on.
There is a distinct contrast between this view and prior approaches to studying attention.
Are we satisfied by simple descriptions or do we need to embrace system-level explanations?
\begin{ack}
We thank Ed Grefenstette, Chris Summerfield and Will Tebbutt for valuable advice.
Freddie Bickford Smith is supported by the Engineering \& Physical Sciences Research Council and Kellogg College, Oxford.
This work was supported by NIH Grant 1P01HD080679, Wellcome Trust Investigator Award WT106931MA and Royal Society Wolfson Fellowship 183029 to Bradley C Love.
\end{ack}

\newpage
\appendix
\section{Task-set properties}
\label{sec:task_set_properties}
Let a task set, $\{ c_1, c_2 \} \subset \{1,2,3,\ldots,1000\}$, be a subset of two of the 1000 ImageNet classes, where each integer class label corresponds to an object (basketball, lemon, zebra and so on).
Due to the diversity of ImageNet classes, any two task sets can differ substantially, most obviously with respect to their visual appearance.
We aim to describe this variation numerically with six task-set properties: clutter, difficulty, scale, {\cnn} similarity, human similarity and semantic similarity.

\paragraph{Clutter}
The amount of irrelevant visual information, or clutter, varies between images.
To measure this, we count the number of distracting cues in each image.
Automating this is made possible by an object-detection model that predicts the locations and identities of objects present in an image, assigning a confidence score to each object it detects.
This model is a Faster R-CNN \citep{ren15} that uses an ImageNet-pretrained Inception ResNet v2 \citep{szegedy17} as its feature extractor and is trained on the Open Images v4 dataset \citep{kuznetsova20}.

Our approach is inspired by \cite{russakovsky15}, who also measured clutter by counting the objects in an image.
But, crucially, whereas they decided to treat all objects equally by using a class-agnostic object detector \citep{alexe12}, we make use of the knowledge that, depending on the subject of an image, some objects are more distracting than others.
This helps us to satisfy three conditions that we identify as important when measuring clutter.

\begin{enumerate}[leftmargin=\enumindent]
\item Each unique object is counted only once.
If this were not the case, an image showing a bunch of bananas would receive a higher clutter score than one that shows a single banana, even if both images contain the same number of distracting cues.
\item Each object's contribution to the clutter score is weighted by the model's confidence that the object is present in the image.
Given ambiguous images and an imperfect model, an object should only count insofar as it actually appears to be in the image.
\item The more consistently an object occurs in images of a given class, the more relevant it is to that class and the less it contributes to the clutter score for an image from that class.
An ice cream should count as clutter in raincoat images more than it does in beach images.
\end{enumerate}

Based on these conditions, we define the clutter score of task set $\{ c_1, c_2 \}$ as the mean of $\textrm{clutter}(c_1)$ and $\textrm{clutter}(c_2)$, with the class-level clutter score given by
\begin{equation*}
\textrm{clutter}(c) = \underset{{x \in \mathcal{X}_c^t}}{\textrm{median}} \left( \sum_{o \in \mathcal{O}_x} \textrm{irrelevance}(c,o) \cdot \textrm{confidence}(x,o) \right)
\end{equation*}
where $\mathcal{X}_c^t$ is the subset of ImageNet training images with label $c$ and $\mathcal{O}_x \subset \{1,2,3,\dots,600\}$ is the subset of Open Images object labels predicted for image $x$.
Conditions 1 and 2 are satisfied by summing the object-recognition model's maximal confidence for each unique object, $o$, in each image, $x$.
Condition 3 is satisfied by including an irrelevance score for image label $c$ and object label $o$, based on a cooccurrence matrix, $C$:
\begin{equation*}
\textrm{irrelevance}(c,o) = 1 - \hat{C}_{c,o} \qquad \qquad C_{c,o} = \sum_{x \in \mathcal{X}_c^t} \sum_{o \in \mathcal{O}_x} \textrm{confidence}(x,o)
\end{equation*}
where $\hat{C}$ is a row-normalised version of $C$.
According to the irrelevance score, the less frequently object $o$ occurs in images of class $c$, the less relevant it is to class $c$.

\paragraph{Difficulty}
Whereas images from some ImageNet classes can be classified with perfect accuracy by modern {\cnn}s, those from others are frequently misclassified \citep{russakovsky15}.
In other words, classes vary in difficulty.
We define the difficulty score of task set $\{ c_1, c_2 \}$ as the mean of $\textrm{difficulty}(c_1)$ and $\textrm{difficulty}(c_2)$, with the difficulty score of a class being the error rate of our baseline model (Section \ref{sec:model}) on images from that class:
\begin{equation*}
\textrm{difficulty}(c) = \frac{1}{|\mathcal{X}_c^v|} \sum_{x \in \mathcal{X}_c^v} \mathbb{I}\left(\arg \max_{c'}(p(c' \mid x, a_0)) \neq c\right)
\end{equation*}
where $\mathcal{X}_c^v$ is the subset of ImageNet validation images with label $c$, $\mathbb{I}$ is an indicator function and $p(c \mid x, a_0)$ is the output of the baseline model.

\begin{figure}
\centering
\includegraphics[width=0.7\textwidth]{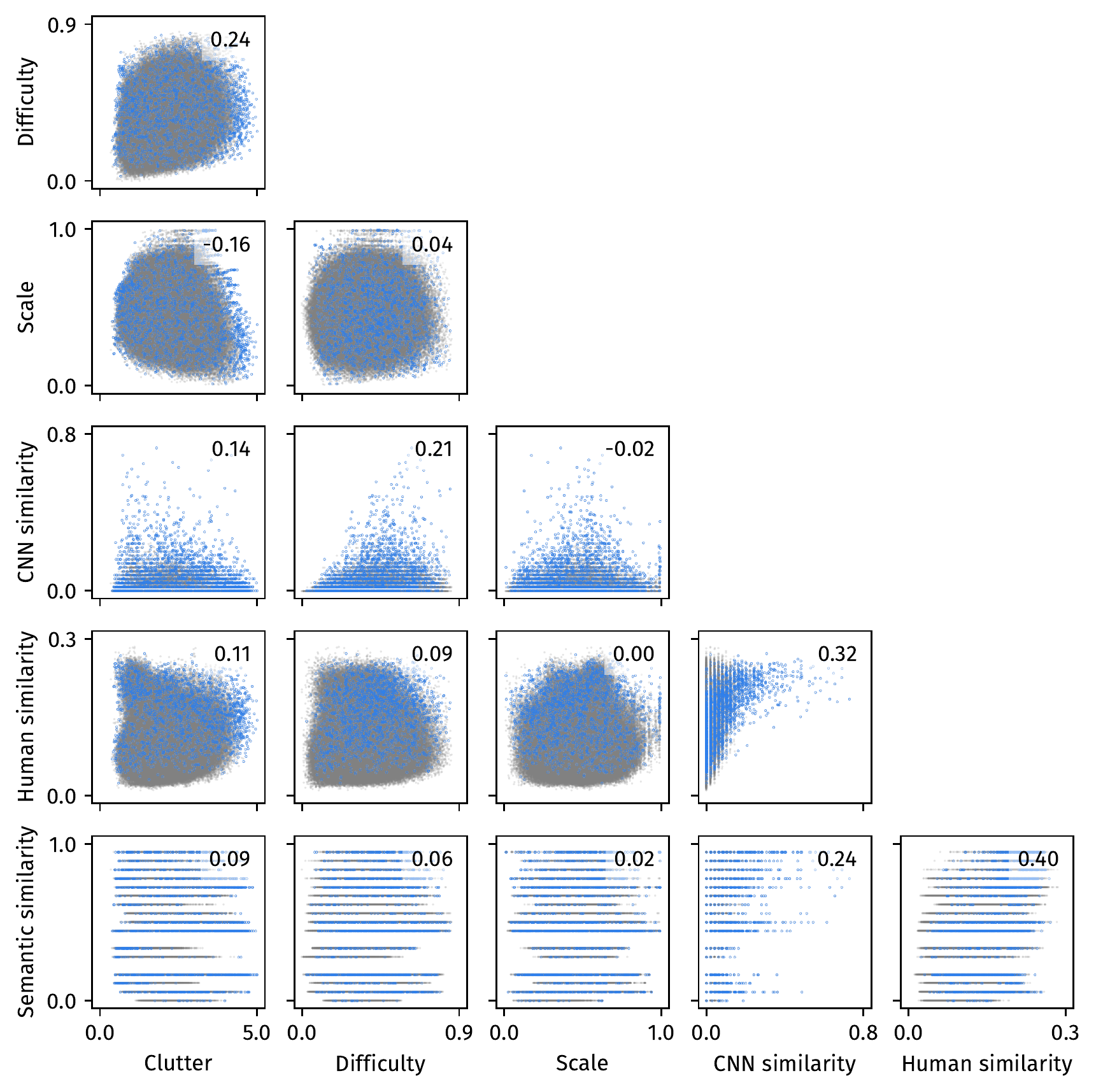}
\caption{
Exploring combinations of task-set properties.
Enumerating all 499,500 possible task sets (pairs of ImageNet classes) and computing their properties gives a set of candidate combinations of task-set properties (grey).
To explore a diverse collection of task sets, we use a quasi-random sequence to identify spaced-out target locations, then for each target location find the task set whose properties (blue) are closest to the target.
In the upper-right corner of each plot, we show the Spearman correlation between the corresponding pair of task-set properties, computed using the grey points in the plot.
}
\label{fig:task_set_properties}
\reducespace
\end{figure}

\paragraph{Scale}
Images vary with respect to scale, the size of the object they portray relative to the size of the whole image.
Scale can be estimated using the ImageNet examples for which human annotators have drawn a bounding box around the subject of an image (or multiple boxes if the subject appears more than once).
We define the scale score of task set $\{ c_1, c_2 \}$ as the mean of $\textrm{scale}(c_1)$ and $\textrm{scale}(c_2)$, where the class-level scale score is the median proportion of an image covered by its subject:
\begin{equation*}
\textrm{scale}(c) = \underset{x \in \mathcal{X}_c^b}{\textrm{median}} \left( \frac{\textrm{area}(\mathcal{B}_x)}{\textrm{area}(x)} \right)
\end{equation*}
where $\mathcal{X}_c^b$ is the subset of ImageNet training images with label $c$ and with a non-empty set of bounding-box annotations, $\mathcal{B}_x$, and area is measured in pixels.
In cases where there are multiple bounding boxes in $\mathcal{B}_x$, we compute the area of their union.
This ensures scale is large for images where multiple small instances of the subject together cover much of the image.

\paragraph{{\cnn} similarity}
Some pairs of ImageNet classes are more visually similar than others.
One way to capture this quantitatively is to look at the mistakes that a {\cnn} makes when classifying images.
Using the ImageNet validation set, we compute the confusion matrix, $D$, of our baseline model (Section \ref{sec:model}), where $D_{i,j}$ denotes how often the predicted class is $j$ when the true class is $i$.
We convert this to a symmetric similarity matrix, $S$ \citep{klein70}:
\begin{equation*}
S_{i,j} = S_{j,i} = \frac{1}{2} \sum_k D_{i,k} + D_{j,k} - |D_{i,k} - D_{j,k}|
\end{equation*}
The {\cnn} similarity of task set $\{ c_1, c_2 \}$ is simply $S_{c_1,c_2}$.
The more often the classes in the task set are confused for each other by the {\cnn}, the greater their similarity.

\paragraph{Human similarity}
When people make comparisons between objects, factors other than visual appearance influence their perception of similarity.
For instance, two objects might be visually unalike yet still be perceived as similar because they are contextually linked: consider cyclists and traffic lights.
To capture a rich notion of similarity that complements the {\cnn}-based measure, we make use of psychological embeddings developed by \cite{roads21}.
These embeddings were inferred from the results of a large-scale experiment in which people were shown a series of query images and for each query asked to identify the most similar images among of a set of candidates.
Having human similarity judgements encoded in the form of embeddings allows us to compute a matrix, $S^h$, that summarises how similarly people perceive each pair of ImageNet classes.
The human similarity of task set $\{ c_1, c_2 \}$ is then $S^h_{c_1,c_2}$.

\paragraph{Semantic similarity}
ImageNet classes correspond to distinct but related concepts.
A pair of classes can be semantically similar (speedboat and canoe), dissimilar (speedboat and bagel) or somewhere in between (speedboat and minibus).
This information is naturally represented in WordNet \citep{miller95}, the semantic hierarchy associated with ImageNet.
The conceptual proximity of two classes can be measured by looking at how far apart the classes are in the hierarchy.
To automate this, we make use of semantic embeddings created by \cite{barz19}.
These embeddings, one for each ImageNet class, were designed such that computing the dot product between a pair of class embeddings yields a similarity score for those classes, corresponding to how close the two classes are in the WordNet hierarchy.
In other words, if $e_1$ and $e_2$ are the embeddings for classes $c_1$ and $c_2$, the semantic similarity of task set $\{ c_1, c_2 \}$ is given by $e_1^\top e_2$.

\begin{figure}
\centering
\begin{subfigure}{0.47\textwidth}
    \includegraphics[width=\textwidth,trim={0.2cm 0.2cm 0.4cm 0}]{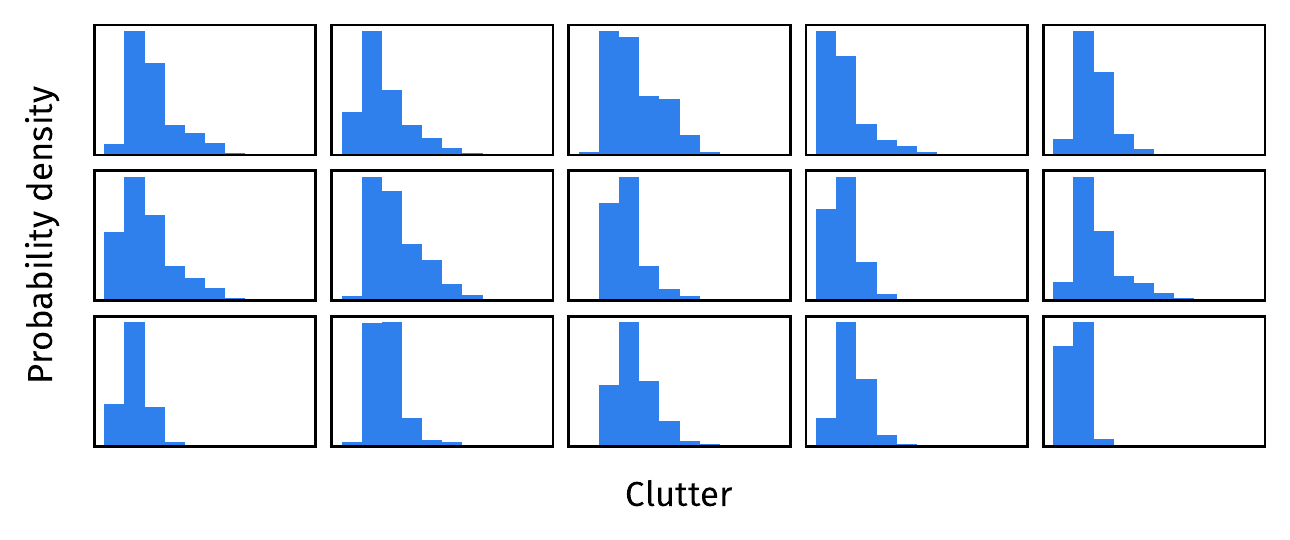}
    \end{subfigure}
\hfill
\begin{subfigure}{0.47\textwidth}
    \includegraphics[width=\textwidth,trim={0.4cm 0.2cm 0.2cm 0}]{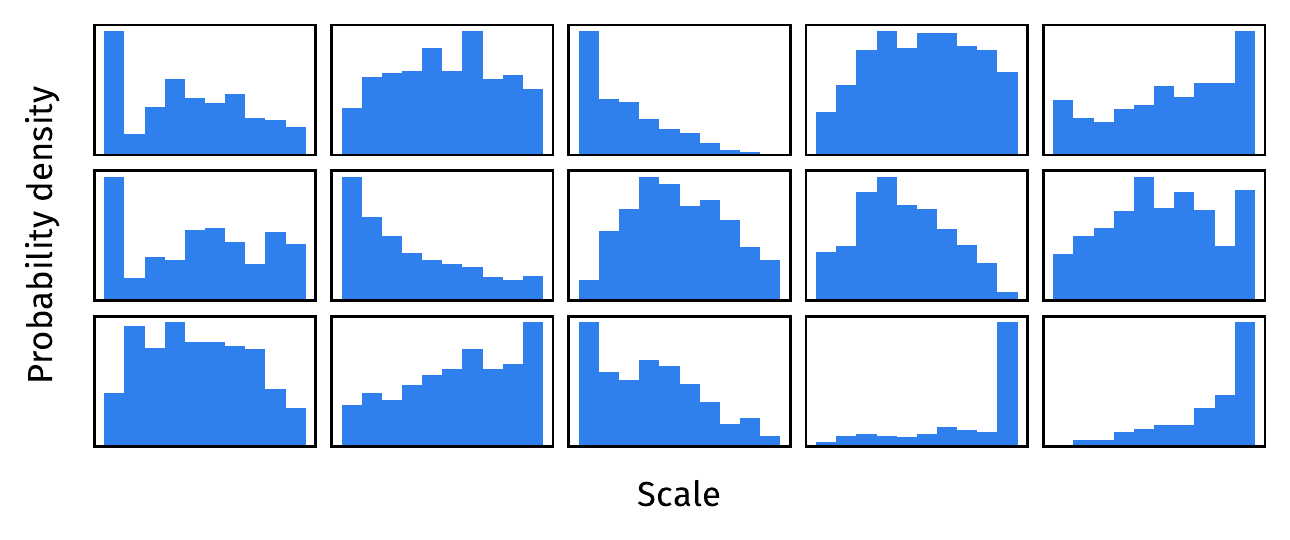}
\end{subfigure}
\caption{
Distributions of image-level clutter and scale scores within randomly sampled ImageNet classes.
Clutter represents how much distracting information is present in an image.
Scale is the size of an image's subject relative to the size of the overall image.
Across each class, there is substantial variance in clutter and scale.
Left: each horizontal axis spans the $[0, 10]$ interval.
Right: each horizontal axis spans the $[0, 1]$ interval.
}
\label{fig:within_class_variation}
\reducespace
\end{figure}

\section{Resources used in this work}
\label{sec:resources}
Our experiment is based on the ImageNet dataset.
The presence of people in ImageNet images raises issues around demographic representation, inappropriate content and privacy.
We strongly support work, such as recent research by \cite{yang21}, that addresses these issues and other ethical concerns associated with ImageNet.

The implementation of our experiment builds on a number of existing software projects (Table \ref{tab:software}).
To run the experiment, we use an internal server equipped with Nvidia GeForce RTX 2080 Ti processors.

\begin{table}[ht]
\centering
\textsf{
{
\scriptsize
\begin{filecontents*}{figures/software_citations.csv}
Project,Citation,License,URL
Jupyter,\cite{kluyver16},BSD (3-clause),\shorturl{jupyter.org}
Matplotlib,\cite{hunter07},PSF (modified),\shorturl{matplotlib.org}
NumPy,\cite{harris20},BSD (3-clause),\shorturl{numpy.org}
Pandas,\cite{mckinney10},BSD (3-clause),\shorturl{pandas.pydata.org}
Python,\cite{vanrossum95},PSF,\shorturl{python.org}
Scikit-learn,\cite{pedregosa11},BSD (3-clause),\shorturl{scikit-learn.org}
SciPy,\cite{virtanen20},BSD (3-clause),\shorturl{scipy.org}
Seaborn,\cite{waskom21},BSD (3-clause),\shorturl{seaborn.pydata.org}
Shapely,\cite{gillies07},BSD,\shorturl{github.com/toblerity/shapely}
sobol\_seq,not available,MIT,\shorturl{github.com/naught101/sobol_seq}
Statsmodels,\cite{seabold10},BSD (3-clause),\shorturl{statsmodels.org}
tqdm,\cite{dacostaluis19},MPL,\shorturl{github.com/tqdm/tqdm}
TensorFlow,\cite{abadi16},Apache 2.0,\shorturl{tensorflow.org}
\end{filecontents*}
\csvreader[
    head to column names,
    tabular=llll,
    table head=\toprule Project & Citation & License & URL \\ \midrule,
    table foot=\bottomrule,
    ]{figures/software_citations.csv}{}{\csvlinetotablerow}
}
}
\caption{
Software projects used in this work.
}
\label{tab:software}
\reducespace
\end{table}
\end{document}